\documentclass[letterpaper, 10 pt, conference]{ieeeconf}
\IEEEoverridecommandlockouts
\usepackage{amssymb}
\usepackage{amsmath}
\usepackage{acro}
\usepackage{pdfpages}
\usepackage{subcaption}
\usepackage{tabularx}
\usepackage{float}
\usepackage{cite}
\usepackage{gensymb}
\usepackage{algorithm, algorithmic}
\usepackage{booktabs}
\usepackage{fancyhdr}

\DeclareMathOperator*{\argmax}{argmax} 
\DeclareAcronym{CPP}{
short = CPP ,
long = coverage path planning
}
\DeclareAcronym{UAV}{
short = UAV ,
long = unmanned aerial vehicle
}
\DeclareAcronym{RL}{
short = RL ,
long = reinforcement learning
}
\DeclareAcronym{DQN}{
short = DQN ,
long = deep Q-network
}
\DeclareAcronym{DDQN}{
short = DDQN ,
long = double deep Q-network
}
\DeclareAcronym{NN}{
short = NN ,
short-plural = NNs ,
long = neural network ,
long-plural = neural networks
}
\DeclareAcronym{FoV}{
short = FoV ,
long = field of view
}
\DeclareAcronym{TD}{
short = TD ,
long = temporal difference
}

\title{\LARGE \bf UAV Coverage Path Planning under Varying Power Constraints using Deep Reinforcement Learning}
\author{Mirco Theile$^{1}$, Harald Bayerlein$^{2}$, Richard Nai$^{1}$, David Gesbert$^{2}$, and  Marco Caccamo$^{1}$%
\thanks{$^{1}$Mirco Theile, Richard Nai, and Marco Caccamo  are  with  the TUM Department of Mechanical Engineering, Technical University of Munich, Germany
        {\tt\small \{mirco.theile, richard.nai, mcaccamo\}@tum.de}}%
\thanks{$^{2}$Harald Bayerlein and David Gesbert are with the Communication Systems Department, \mbox{EURECOM}, Sophia Antipolis, France
        {\tt\small \{harald.bayerlein, david.gesbert\}@eurecom.fr}}%
}

\begin{document}

\maketitle
\thispagestyle{empty}
\pagestyle{empty}

\begin{abstract}
\Ac{CPP} is the task of designing a trajectory that enables a mobile agent to travel over every point of an area of interest. We propose a new method to control an \ac{UAV} carrying a camera on a \ac{CPP} mission with random start positions and multiple options for landing positions in an environment containing no-fly zones. While numerous approaches have been proposed to solve similar CPP problems, we leverage end-to-end \ac{RL} to learn a control policy that generalizes over varying power constraints for the UAV. Despite recent improvements in battery technology, the maximum flying range of small UAVs is still a severe constraint, which is exacerbated by variations in the UAV's power consumption that are hard to predict. By using map-like input channels to feed spatial information through convolutional network layers to the agent, we are able to train a \ac{DDQN} to make control decisions for the \ac{UAV}, balancing limited power budget and coverage goal. The proposed method can be applied to a wide variety of environments and harmonizes complex goal structures with system constraints.
\end{abstract}
\section{Introduction}
Whereas the \ac{CPP} problem for ground-based robotics has already found its way into our everyday life in the form of vacuum cleaning robots \cite{Galceran2013}, autonomous coverage with UAVs, while not yet having attained the same level of prominence, is being considered for a wide range of applications, such as photogrammetry, smart farming and especially disaster management \cite{Cabreira2019}. UAVs can be deployed rapidly to gather initial or continuous survey data of areas hit by natural disasters, or mitigate their consequences. In the aftermath of the 2019-20 Australian bushfire season, wildlife officers inventively used quadcopter drones with infrared sensors to conduct a search-and-rescue operation for koalas affected by the blaze \cite{Gimsey2020}.

As its name suggests, covering all points inside an area of interest with CPP is related to conventional path planning where the goal is to find a path between start and goal positions. In general, CPP aims to cover as much of the target area as possible within given energy or path-length constraints while avoiding obstacles or no-fly zones. Due to the limitations in battery energy density, available power limits mission duration for quadcopter UAVs severely. Finding a CPP control policy that generalizes over varying power constraints and setting a specific movement budget can be seen as a way to model the variations in actual power consumption during the mission, e.g. caused by environmental factors that are hard to predict. Similar to conventional path planning, CPP can usually be reduced to some form of the travelling salesman problem, which is NP-hard \cite{Galceran2013}. Lawn-mowing and milling \cite{Arkin2000} are other examples of closely related problems.

The most recent survey of UAV coverage path planning is given by Cabreira \textit{et al.} \cite{Cabreira2019}. Galceran and Carreras \cite{Galceran2013} provide a survey of general (ground robotics) approaches to CPP. Autonomous UAVs for applications in wireless communications have also sparked a lot of interest recently. Some scenarios, e.g. deep RL trajectory planning for UAVs providing wireless connectivity under battery power constraints, are related to CPP. An overview of UAV applications in wireless communications can be found in \cite{Zeng2019}.

To guarantee complete coverage, most existing CPP approaches split the target area and surrounding free space into cells, by means of exact or approximate cellular decomposition. Choset and Pignon \cite{Choset1998} proposed the classical boustrophedon (``the way of the ox'', back and forth motion) cellular decomposition, an exact decomposition method that guarantees full coverage but offers no bounds on path-length. This algorithm was extended by Mannadiar and Rekleitis \cite{Mannadiar2010} through encoding the cells of the boustrophedon decomposition as a Reeb graph and then constructing the Euler tour that covers every edge in the graph exactly once. Cases where the mobile agent does not have enough power to cover the whole area are not considered. The authors in \cite{Xu2011} adapted this method for use in a non-holonomic, fixed-wing UAV and conducted extensive experimental validation. Two other approaches combining CPP and the travelling salesman problem to find near-optimal solutions for coverage of target regions enclosed by non-target areas are proposed by the authors in \cite{Xie2018}: grid-based and dynamic programming-based, respectively. Both approaches suffer from exponential increase in time complexity with the number of target regions and do not consider obstacles or UAV power limitations.

Non-standard approaches have made use of neural networks (NNs) before. The authors in \cite{Yang2004} design a network of neurons with only lateral connections that each represent one grid cell in a cleaning robot's non-stationary 2D environment. The path planning is directly based on the neural network's activity landscape, which is computationally simple and can support changing environments, but does not take path-length or power constraints into account.

Reinforcement learning with deep neural networks has only recently started to be considered for UAV path planning. Maciel-Pearson \textit{et al.} \cite{MacielPearson2019} proposed a method using an extended double deep Q-network (EDDQN) to explore and navigate from a start to a goal position in outdoor environments by combining map and camera information from the drone. Their approach is focused on obstacle avoidance under changing weather conditions. The authors in  \cite{Piciarelli2019} investigate the CPP-related drone patrolling problem where a UAV patrols an area optimizing the relevance of its observations through the use of a single-channel relevance map fed into a convolutional layer of a DDQN agent. However, there is no consideration for power constraints and the relevance map is preprocessed showing only local information. To the best of our knowledge, deep RL has not been considered for UAV control in coverage path planning under power constraints before.

The main contributions of this paper are the following:
\begin{itemize}
    \item Introduction of a novel UAV control method for coverage path planning based on double deep Q-learning;
    \item The usage of map-like channels to feed spatial information into convolutional network layers of the agent;
    \item Learning a control policy that generalizes over random start positions and varying power constraints and decides between multiple landing positions.
\end{itemize}

The remainder of this paper is organized as follows: Section \ref{sec:problem} introduces the CPP problem formulation, Section \ref{sec:methodology} describes our DDQN learning approach and in Section \ref{sec:experiments} follow simulation results and their discussion. We conclude the paper with a summary and outlook onto future work in Section \ref{sec:conclusion}.
\section{Problem Formulation}
\label{sec:problem}

\subsection{Setup}
\begin{figure}
    \vspace{5pt}
    \centering
    \includegraphics[width=0.9\columnwidth]{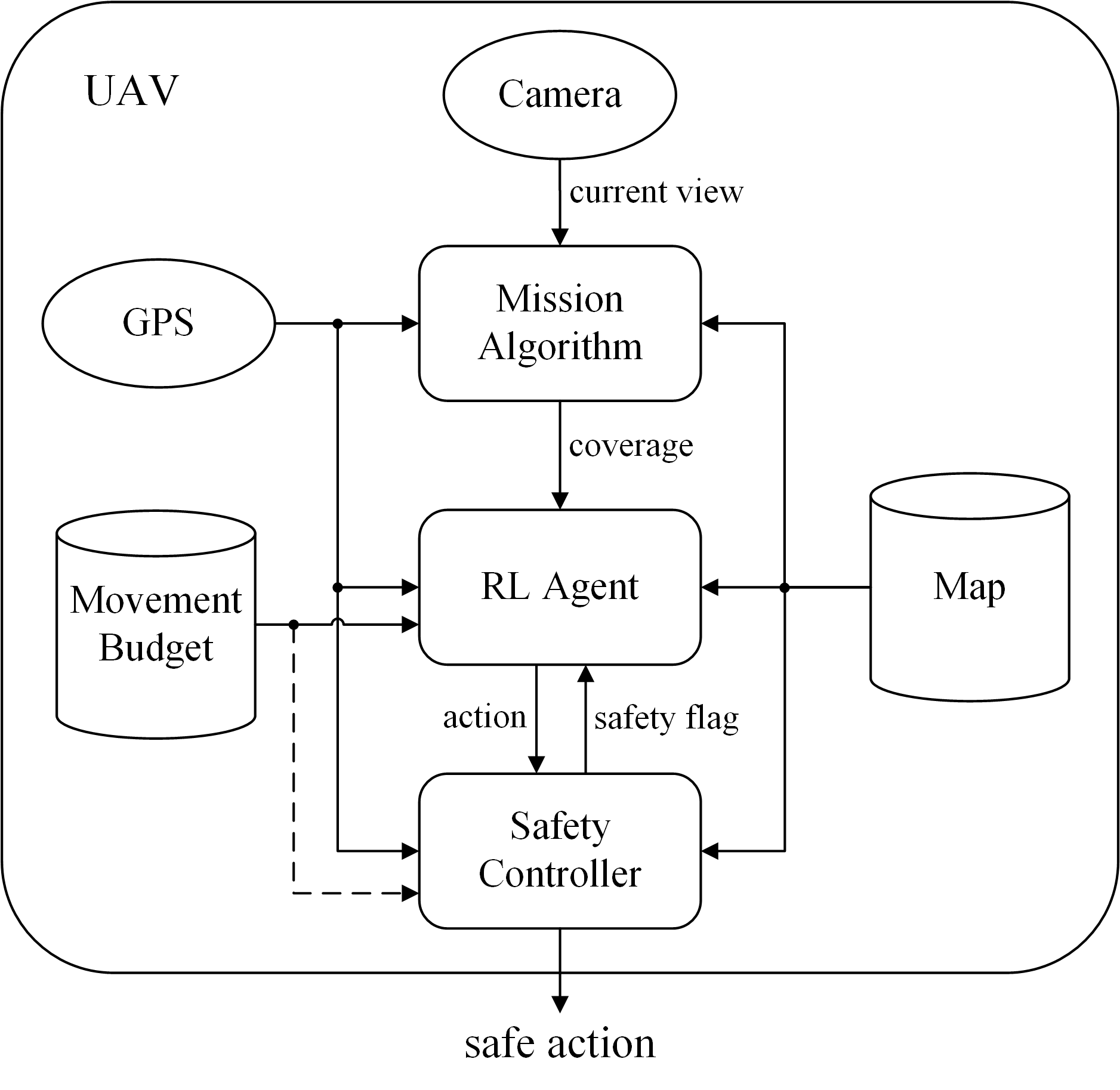}
    \caption{System-level diagram depicting sensor and software components on the UAV during a coverage mission.}
    \vspace{-5pt}
    \label{fig:system_model}
\end{figure}

The sensors of the UAV forming the input of the reinforcement learning agent are depicted in Figure \ref{fig:system_model}: camera and GPS receiver. The camera gives a periodic frame of the current coverage view and the GPS yields the drone's position. Power constraints determined by external factors are modelled as a movement budget for the drone that is fixed at mission start. Two additional software components are running on the UAV. The first is the mission algorithm which is responsible for the analysis of the camera data. We assume that any mission algorithm can give feedback on the area that was already covered. The second component is a safety controller that evaluates the proposed action of the agent and accepts or rejects it based on the safety constraints (entering into no-fly zones or landing in unsuitable areas). Note that the safety controller does not assist the agent in finding the landing area. The last component is a map which is provided by the operator on the ground. While this map could be dynamic throughout the mission, we focus on static maps for the duration of one mission in this paper.

\subsection{3-Channel Map}
The coverage problem to be solved can be represented by a two dimensional grid map with three channels. Each cell in the grid represents a square area of the coverage region. The three channels describe starting and landing zones, target zones, and no-fly zones. The start and landing zones are areas the agent can start from and land on after finishing a coverage path. Target zones have to be covered at least once by the field of view (FoV) of the UAV's camera. No-fly zones represent areas which the drone is prohibited from entering. Note that it is possible that a cell is declared as none, or more than one of these zones, with the exception that starting and landing zones can not be no-fly zones at the same time.

\subsection{Markov Decision Process}
In order to solve the described coverage path planning problem with reinforcement learning, it is converted into a Markov decision process (MDP). An MDP is described by the tuple $(\mathcal{S}, \mathcal{A}, R, P)$, with the set of possible states $\mathcal{S}$, the set of possible actions $\mathcal{A}$, the reward function $R$, and the deterministic state transition function $P : \mathcal{S} \times \mathcal{A} \mapsto S$. 

In a $N\times N$ grid, the state space $\mathcal{S}$ has the following dimensions:
\begin{equation*}
    \mathcal{S} = \underbrace{\mathbb{B}^{N\times N \times 3}}_{\text{Map}}\times\underbrace{\mathbb{B}^{N\times N}}_{\text{Coverage}}\times \underbrace{\mathbb{R}^2}_{\text{Position}}\times \underbrace{\mathbb{N}}_{\text{Movement Budget}}\times\underbrace{\mathbb{B}}_{\text{Safety Flag}}
\end{equation*}
where $\mathbb{B}$ is the Boolean domain $\{0,1\}$. The action space $\mathcal{A}$ contains the following five actions:
\begin{equation*}
    \mathcal{A} = \{\text{north}, \text{east}, \text{south}, \text{west}, \text{land}\}
\end{equation*}
The reward function $R : \mathcal{S} \times \mathcal{A} \mapsto \mathbb{R}$, mapping the current state $s \in \mathcal{S}$ and current action $a \in \mathcal{A}$ to a real-valued reward, consists of multiple components:
\begin{itemize}
    \item $r_{cov}$ \textit{(positive)} coverage reward for each target cell that is covered by the UAV's field of view for the first time;
    \item $r_{sc}$ \textit{(negative)} safety penalty in case the safety controller (SC) rejects the agent's proposed action;
    \item $r_{mov}$ \textit{(negative)} constant movement penalty that is applied for every unit of the movement budget the UAV uses
    \item $r_{crash}$ \textit{(negative)} penalty in case the UAV runs out of movement budget without having safely landed in a landing zone.
\end{itemize}
\section{Methodology}
\label{sec:methodology}

\subsection{Q-Learning}
Reinforcement learning, in general, proceeds in a cycle of interactions between an agent and its environment. At time $t$, the agent observes a state $s_t \in \mathcal{S}$, performs an action $a_t \in \mathcal{A}$ and subsequently receives a reward $r(s_t, a_t) \in \mathbb{R}$. The time index is then incremented and the environment propagates the agent to a new state $s_{t+1}$, from where the cycle restarts. The goal of the agent is to maximize the discounted cumulative return $R_t$ from the current state up to a terminal state at time $T$. It is given as
\begin{equation}
    R_t = \sum_{k=t}^T \gamma^{k-t} r(s_k,a_k).
\end{equation}
with $\gamma \in [0, 1]$ being the discount factor, balancing the importance of immediate and future rewards. The return is maximized by adapting the agent's behavioral policy $\pi$. The policy can be deterministic with $\pi(s)$ such that $\pi : \mathcal{S}\mapsto\mathcal{A}$, or probabilistic with $\pi(a|s)$ such that $\pi : \mathcal{S}\times\mathcal{A}\mapsto\mathbb{R}$, yielding a probability distribution over the action space for each $s\in\mathcal{S}$.

To find a policy which maximizes the return, we utilize Q-learning, a model-free reinforcement learning approach. It is based on learning the state-action-value function, or \emph{Q-function} $Q:\mathcal{S}\times\mathcal{A}\mapsto\mathbb{R}$, defined as
\begin{equation}
    Q^\pi(s,a) = \mathbb{E}_{\pi} \left[R_t | s_t = s, a_t = a\right]
    \label{eq:q}.
\end{equation} 
Q-learning relies on iteratively updating the current knowledge of the Q-function. When the optimal Q-function is known, it is easy to construct an optimal policy by taking actions that maximize the Q-function. For convenience, $s_t$ and $a_t$ are abbreviated to $s$ and $a$ and $s_{t+1}$ and $a_{t+1}$ to $s^\prime$ and $a^\prime$ in the following.

\subsection{Deep Q-Learning}
The Q-function from \eqref{eq:q} can be represented through a table of Q-values with the dimension $\mathcal{S}\times\mathcal{A}$. This is not feasible for large state or action spaces, but it is possible to approximate the Q-function by a neural network in those cases. A deep Q-network (DQN) parameterizing the Q-function with the parameter vector $\theta$ is trained to minimize the expected temporal difference (TD) error given by
\begin{equation}
    L(\theta) = \mathbb{E}_\pi[(Q_\theta(s,a) - Y(s,a,s^\prime))^2]
\end{equation}
with the target value
\begin{equation}
    Y(s,a,s^\prime) = r(s,a) + \gamma \max_{a^\prime}Q_\theta(s^\prime,a^\prime).
\end{equation}
While a DQN is significantly more data efficient compared to a Q-table due to its ability to generalize, the \mbox{\textit{deadly triad} \cite{Sutton2018}} of function approximation, bootstrapping and off-policy training can make its training unstable and cause divergence. In 2015, Mnih \textit{et al.} \cite{Mnih2015} presented a methodology to stabilize the DQN learning process. Their training approach makes use of an experience replay memory $\mathcal{D}$ which stores experience tuples $(s, a, r, s^\prime)$ collected by the agent during each interaction with the environment. Training the agent on uniformly sampled batches from the replay memory decorrelates the individual samples and rephrases the TD-error as
\begin{equation}
    L^{\text{DQN}}(\theta) = \mathbb{E}_{s,a,s^\prime \sim \mathcal{D}}[(Q_\theta(s,a) - Y^{\text{DQN}}(s,a,s^\prime))^2].
    \label{eq:loss}
\end{equation}
Additionally, Mnih \textit{et al.} used a separate target network for the estimation of the next maximum Q-value changing the target value to
\begin{equation}
    Y^{\text{DQN}}(s,a,s^\prime) = r(s,a) + \gamma \max_{a^\prime}Q_{\bar{\theta}}\left(s^\prime, a^\prime\right)
    \label{eq:td_target}
\end{equation}
with $\bar{\theta}$ representing the parameters of the target network. The parameters of the target network $\bar{\theta}$ can either be updated as a periodic hard copy of $\theta$ or as a soft update with
\begin{equation}
    \bar{\theta} \leftarrow (1-\tau)\bar{\theta} + \tau\theta
    \label{eq:soft_update}
\end{equation}
after each update of $\theta$. $\tau \in [0,1]$ is the update factor determining the adaptation pace. The combination of replay memory and target network separation to stabilize the training process laid the groundwork for the rise in popularity of DQN methods.

An additional improvement was proposed by Van Hasselt \textit{et al.} \cite{VanHasselt2016}, who showed that under certain conditions, action values in \eqref{eq:td_target} get overestimated. To solve this issue, the double deep Q-network (DDQN) was introduced. The target value is then given by
\begin{equation}
    Y^{\text{DDQN}}(s,a,s^\prime) = r(s,a) + \gamma Q_{\bar{\theta}}(s^\prime, \argmax_{a^\prime}Q_{\theta}(s^\prime, a^\prime))
    \label{eq:ddqn}
\end{equation}
and the corresponding loss function
\begin{equation}
    L^{\text{DDQN}}(\theta) = \mathbb{E}_{s,a,s^\prime \sim \mathcal{D}}[(Q_\theta(s,a) - Y^{\text{DDQN}}(s,a,s^\prime))^2],
    \label{eq:loss_ddqn}
\end{equation}
in which the overestimation of action values is reduced by selecting the best action using $\theta$ but estimating the value of that action using $\bar{\theta}$. When calculating $\nabla_\theta L^{\text{DDQN}}(\theta)$ the target value is taken as is, hence, the back-propagating gradient is stopped before $Y^{\text{DDQN}}(s,a,s^\prime)$.

\subsection{Neural Network Model and Training Procedure}
\begin{figure*}
    \vspace{5pt}
    \centering
    \includegraphics[width=0.95\textwidth]{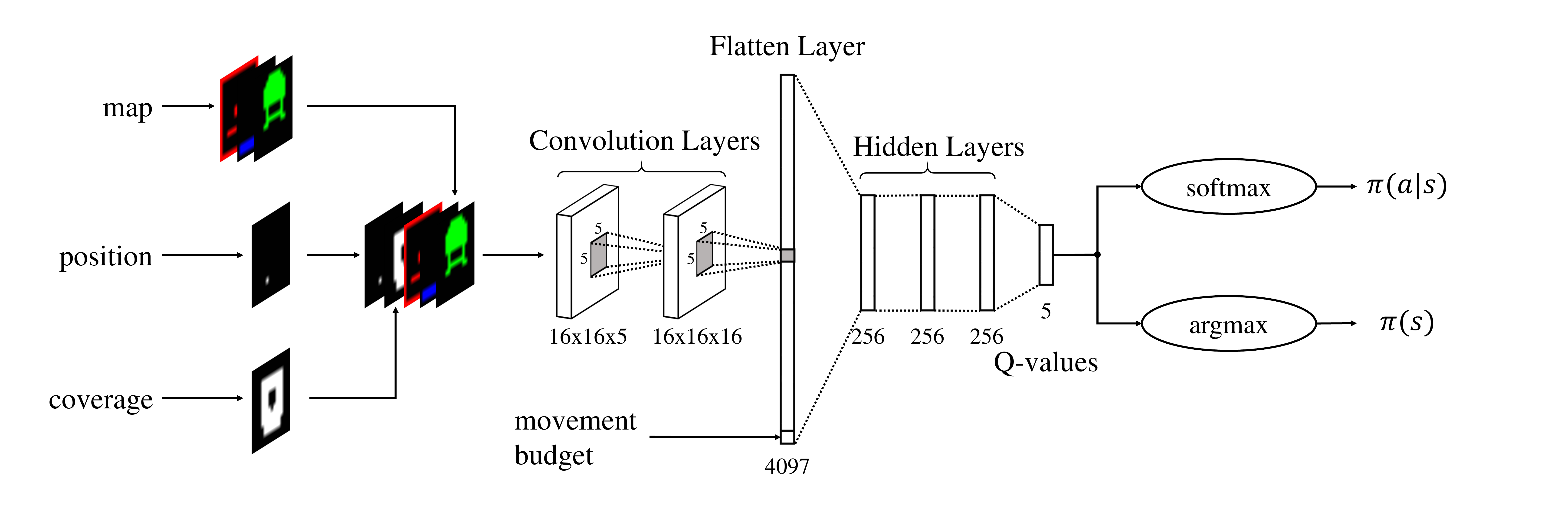}
    \caption{Neural network structure for the reinforcement learning agent.}
    \label{fig:network}
    \vspace{-5pt}
\end{figure*}

The DQN solving the MDP from Section \ref{sec:problem} consists of convolutional and fully-connected layers. It is visualized in Figure \ref{fig:network}. The UAV's own position is converted to a 2D one-hot representation, i.e. the encoding of the occupied cell inside the whole grid. With the position encoded in this way, it can be stacked with the three-channel map and the coverage grid to form the five-channel input of the network's convolutional layers. The kernels of the convolutional layers are then able to form direct spatial connections between the current position and nearby cells. The remaining movement budget is fed into the network after the convolutional layers.

The convolutional layers are padded so that their output shape remains the same as their input shape. All layers are zero-padded for all channels, with the exception of the first layer's no-fly zone channel, which is one-padded. This is an explicit representation of the no-fly zone surrounding the mission grid. The rectified linear unit (ReLU) is chosen as activation function for the convolutional layers. The last layer of the convolutional network is flattened and concatenated with the movement budget input. Fully-connected layers with ReLU activation are attached to this flatten layer.  

The last fully-connected layer is of size $|\mathcal{A}|$ and has no activation function. It directly represents the Q-values for each action given the input state. Choosing the $\argmax$ of the Q-values is called the greedy policy and exploits already learned knowledge. The greedy policy given by
\begin{equation}
    \pi(s) = \argmax_{a\in\mathcal{A}}Q_\theta(s,a)
    \label{eq:greedy}
\end{equation}
is deterministic and used when evaluating the agent's learning progress. During training, the sampled soft-max policy for exploration of the state and action space is used instead. It is given by 
\begin{equation}
    \pi(a_i|s) = \frac{\mathrm{e}^{Q_\theta(s,a_i)/\beta}}{\sum_{\forall a_j \in \mathcal{A}}\mathrm{e}^{Q_\theta(s,a_j)/\beta}}
    \label{eq:softmax}
\end{equation}
with the temperature parameter $\beta \in \mathbb{R}$ scaling the balance of exploration versus exploitation. When $\beta$ is increased so does exploration. The limit $\beta\rightarrow 0$ of the soft-max policy \eqref{eq:softmax} is the greedy policy \eqref{eq:greedy}. The soft-max policy was chosen over the $\epsilon$-greedy policy because it offers variable exploration based on the relative difference of Q-values and does not depend on the number of training steps or episodes. This appeared to be beneficial for this particular problem.

\begin{algorithm}
\caption{DDQN training for coverage path planning}
\label{alg:alg}
\begin{algorithmic}[1]
\renewcommand{\algorithmicrequire}{}
\REQUIRE Initialize $\mathcal{D}$, initialize $\theta$ randomly, $\bar{\theta}\leftarrow \theta$

\FOR {$n = 0$ to $N_{max}$}
	\STATE Initialize state $s_0$ with random starting position and sample initial movement budget $b_0$ uniformly from $\mathcal{B}$ 
	
	\WHILE {$b > 0$ and not landed} \vspace{2pt}
		\STATE Sample $a$ according to \eqref{eq:softmax}
		\STATE Observe $r$, $s^\prime$
		\STATE Store $(s, a, r, s^\prime)$ in $\mathcal{D}$
		\FOR {$i = 1$ to $m$}
			\STATE Sample $(s_i,a_i,r_i,s^\prime_i)$ uniformly from $\mathcal{D}$
			\STATE $Y_i=\begin{cases} r_i, & \text{if $s^\prime_i$ terminal}\\ \text{according to \eqref{eq:ddqn}}, & \text{otherwise} \end{cases}$
			\STATE Compute loss $L_i (\theta)$ according to \eqref{eq:loss_ddqn}
		\ENDFOR
		\STATE Update $\theta$ with gradient loss $\frac{1}{m}\sum_{i=1}^{m} L_i(\theta)$
	    \STATE Soft update of $\bar{\theta}$ according to \eqref{eq:soft_update}
		\STATE $b = b - 1$
	\ENDWHILE
	
\ENDFOR
\end{algorithmic} 
\end{algorithm}

Algorithm \ref{alg:alg} describes the training procedure for the double deep Q-network in more detail. After replay memory and network parameters are initialized, a new training episode begins with resetting the state, choosing a random UAV starting position and random movement budget $b_0 \in \mathcal{B}$. The episode continues as long as the movement budget is greater than zero and the UAV has not landed. A new action $a \in \mathcal{A}$ is chosen according to \eqref{eq:softmax} and the subsequent experience stored in the replay memory buffer $\mathcal{D}$.

Sampling a minibatch of size $m$ from the replay memory, the primary network parameters $\theta$ are updated by performing a gradient step using the Adam optimizer. Subsequently, the target network parameters $\bar{\theta}$ are updated using the soft update \eqref{eq:soft_update} and the movement budget is decremented. The episode ends when either the drone lands or the movement budget decreases to zero. Then, a new episode starts unless the maximum number of episodes $N_{max}$ is reached. The hyperparameters that were used during training are listed in Table \ref{table:parameters}.

\begin{table}

\vspace{5pt}
\center
\small
\begin{tabular*}{\columnwidth}{ccc}
\toprule[1.5pt]
Parameter & Value & Description\\
\midrule
$|\theta|$ & 1,190,389 & number of trainable parameters\\
$|\mathcal{D}|$ & 50,000 & replay memory buffer size\\
$N_{max}$ & 10,000 & maximum number of training episodes\\
$\beta$ & 0.1 & temperature parameter \eqref{eq:softmax}\\
$m$ & 128 & minibatch size\\
$\gamma$ & 0.95 & discount factor for target value in \eqref{eq:ddqn}\\
$\tau$ & 0.005 & target network update factor \eqref{eq:soft_update}\\
\bottomrule[1.5pt]
\end{tabular*}
\caption{Hyperparameters for DDQN training.}
\label{table:parameters}

\vspace{-15pt}
\end{table}
\section{Experiments}
\label{sec:experiments}

\subsection{Simulation Setup}

The agent can move in a two dimensional grid through action commands in $\mathcal{A}$ if accepted by the safety controller. Each action, no matter if accepted or rejected, consumes one unit of movement budget since energy is spent during hovering as well. The agent's initial state $s_0 \in \mathcal{S}$ consists of a fixed map, a zero-initialized coverage grid and a position, which is uniformly sampled from the starting and landing zone of the map. Additionally, the initial movement budget is uniformly sampled from a movement budget range $\mathcal{B}$, which is set to 25-75 for the purpose of this evaluation. The value of the safety flag in $s_0$ is initialized to zero and the UAV's camera field of view (FoV) is set to a fixed 3-by-3-cell area centered underneath the agent. After each step the mission algorithm marks the FoV as seen in the coverage grid map.

Three evaluation scenarios were chosen, each with a unique problem for the agent to solve. Map A depicted in Figure \ref{fig:maps} (a)-(c) has a large starting and landing zone, which yields high variation during training. Additionally, the shape of the target area is challenging to cover. The difficulty of map B in Figure \ref{fig:maps} (d)-(f) lies in the yellow area that is marked as target zone, but also marked as a no-fly zone, and therefore must be covered by flying adjacent to it. Map C, in Figure \ref{fig:maps} (g)-(i) with a narrow passage between no-fly zones, while easy to cover is very difficult for training as discussed later.

\subsection{Evaluation}

\begin{figure*}
\begin{minipage}{0.6\textwidth}
    \vspace{5pt}
    \newcommand\factor{0.31}
    \newcommand\x{\textwidth}
    \newcommand\vsp{3pt}
    \begin{subfigure}{\factor\x}
        \includegraphics[width=\textwidth]{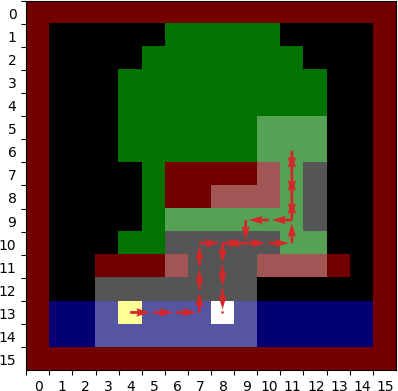}
        \caption{25/25 movement}
    \end{subfigure}
    \begin{subfigure}{\factor\x}
        \includegraphics[width=\textwidth]{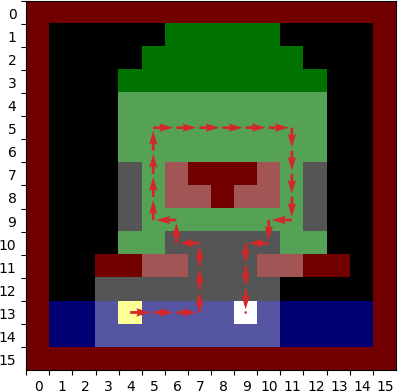}
        \caption{30/30 movement}
    \end{subfigure}
    \begin{subfigure}{\factor\x}
        \includegraphics[width=\textwidth]{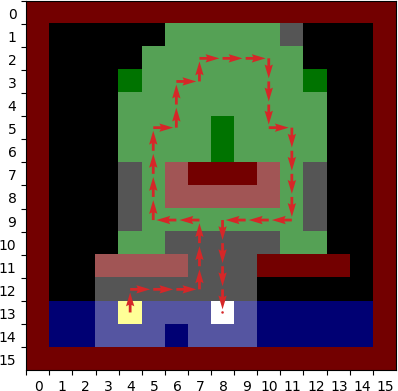}
        \caption{37/37 movement}
    \end{subfigure}
    \hfill\break
    \begin{subfigure}{\factor\x}
        \vspace{\vsp}
        \includegraphics[width=\textwidth]{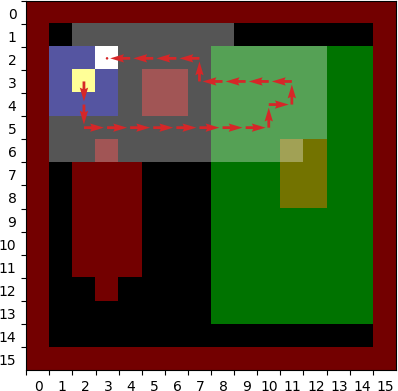}
        \caption{23/25 movement}
    \end{subfigure}
    \begin{subfigure}{\factor\x}
        \vspace{\vsp}
        \includegraphics[width=\textwidth]{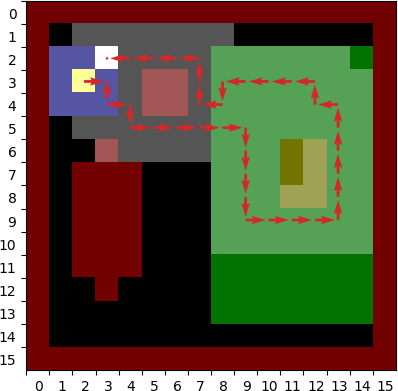}
        \caption{37/40 movement}
    \end{subfigure}
    \begin{subfigure}{\factor\x}
        \vspace{\vsp}
        \includegraphics[width=\textwidth]{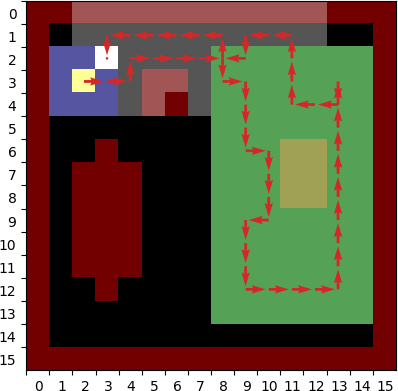}
        \caption{51/60 movement}
    \end{subfigure}
    \hfill\break
    \begin{subfigure}{\factor\x}
        \vspace{\vsp}
        \includegraphics[width=\textwidth]{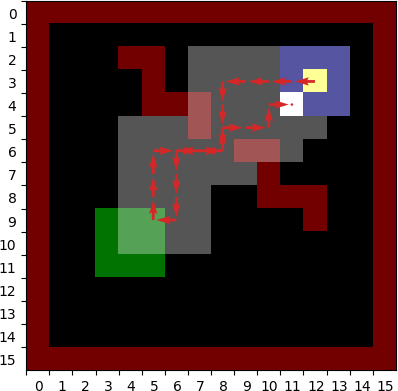}
        \caption{25/25 movement}
    \end{subfigure}
    \begin{subfigure}{\factor\x}
        \vspace{\vsp}
        \includegraphics[width=\textwidth]{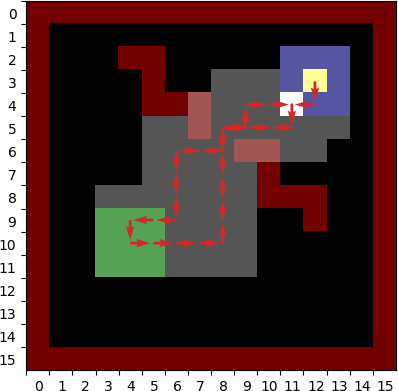}
        \caption{29/29 movement}
    \end{subfigure}
    \begin{subfigure}{\factor\x}
        \vspace{\vsp}
        \includegraphics[width=\textwidth]{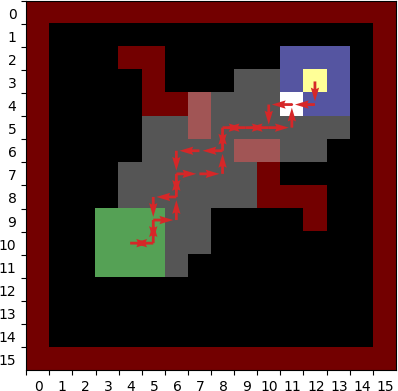}
        \caption{29/40 movement}
    \end{subfigure}\hfill
    \caption{Coverage plots for three different maps (map A: (a)-(c), map B: (d)-(f), map C: (g)-(i)) with three different movement budgets each; red, blue, and green are no-fly zones, starting/landing zones, and target zones, respectively; the red arrows describe the trajectory and the yellow and white cell describe start and landing position, respectively; lighter cells were covered by the agent's FoV.}
    \label{fig:maps}
    \vspace{-5pt}
\end{minipage}
\begin{minipage}{0.02\textwidth}
\hfill
\end{minipage}
\begin{minipage}{0.35\textwidth}
    \begin{minipage}{\textwidth}
        \centering
        \small
        \begin{tabular}{c|c|c|c}
           & Map A  & Map B & Map C \\ \hline
         Landing ratio  &    99.37\% &  99.78\% & 98.26\%
        \end{tabular}
        \captionof{table}{Landing ratio for each map evaluated on the full range of movement budgets and possible starting positions.}
        \label{tab:landing_ratio}
    \end{minipage}\hfill\break
    \begin{minipage}{\textwidth}
        \includegraphics[width=\textwidth]{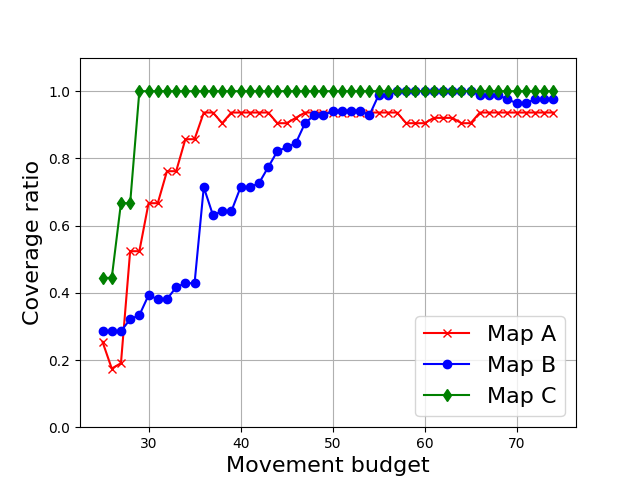}
        \caption{Coverage ratio with varying movement budget for the three maps.}
        \label{fig:coverage}
    \end{minipage}\hfill\break
    \begin{minipage}{\textwidth}
        \includegraphics[width=\textwidth]{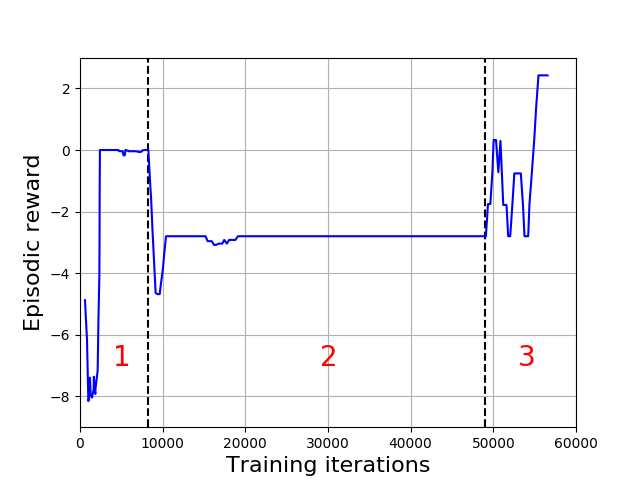}
        \caption{Training process of an agent trained on map C with dashed lines indicating training phase transitions.}
        \label{fig:training}
    \end{minipage}\hfill
\end{minipage}
\vspace{-10pt}
\end{figure*}

After being trained on their respective scenario with varying movement budgets and varying starting positions under the exploration policy $\pi(a|s)$ from \eqref{eq:softmax}, the agents are evaluated under their exploitation policy $\pi(s)$ from \eqref{eq:greedy}. Their performance during coverage for the full movement budget range and all possible starting positions is evaluated. The performance is described through Figures \ref{fig:maps} and \ref{fig:coverage} and Table \ref{tab:landing_ratio}.

The agents' ability to plan a trajectory that ends with a safe landing inside the landing zone over the full movement budget range and starting at each possible position is evaluated through Table \ref{tab:landing_ratio}, showing the ratio of landing for all scenario variations. Despite the agent's good landing performance, the safety controller's capabilities on a real-world UAV would likely be extended to force navigation to the closest landing zone in the rare cases when the RL agent misses the right moment to return.

To evaluate the impact of movement budget on the achieved coverage ratio, the starting position was fixed. For the selected starting positions the agents successfully landed after completing a trajectory for each movement budget. Figure \ref{fig:coverage} shows the coverage ratio of each agent with respect to initial movement budget. Selected trajectories for each map under three different movement budgets are depicted in Figure \ref{fig:maps}. Whereas the movement budget is increasing from left to right, the agent does not necessarily utilize the whole allocated budget if it determines that there is a risk of not returning to the landing area in time or the coverage goal is already fulfilled. It can be seen that the agent finds a trajectory balancing the goals of safe landing and maximum coverage ratio.

Figure \ref{fig:training} shows the training process of an agent on map C. The curve describes the cumulative reward of the exploitation strategy when evaluated during the training process. Three major phases appear during the training process, which are highlighted in the graph. In phase one the agent learns to land safely, but does not venture far enough from the landing zone to find the target zone. When transitioning to phase two, the agent discovers the target zone, yielding high immediate reward. Due to the focus on mid-term reward through the choice of discount factor $\gamma$, this strategy represents a local optimum. In phase three the agent discovers the path back to the landing zone, avoiding the crashing penalty $r_{crash}$. After refining the trajectory, the agent finds the optimal path at the end of phase three. The phase transitions are highly dependent on the exploration strategy. Soft-max exploration appeared to be more effective than the $\epsilon$-greedy policy to guide these transitions. The basic pattern of this incremental learning process is also visible when applied to other maps, albeit with less pronounced transitions due to bigger variations in coverage ratios.

\section{Conclusion}
\label{sec:conclusion}
We have introduced a new deep reinforcement learning approach for the longstanding problem of coverage path planning. While existing approaches might offer guarantees on the (near)-optimality of their solutions, the case where available power constrains the path planning is usually not considered. By feeding spatial information through map-like input channels to the agent, we train a double deep Q-network to learn a UAV control policy that generalizes over varying starting positions and varying power constraints modelled as movement budgets. Using this method, we observed an incremental learning process that successfully balances safe landing and coverage of the target area on three different environments, each with unique challenges.

In the future we will investigate the possibilities of transfer learning for this set of problems. At first we will train the agents on a curriculum of problems based on individual map channels to further accelerate the training process described in this work. From there we will examine approaches to transfer the two dimensional grid agent to higher dimensions and dynamics, e.g. adding altitude and heading. To this effect, it might be beneficial to investigate other RL techniques such as actor-critic methods and policy optimization. The proposed approach can also be seen as an initial step for handling variable power consumption in a real-world scenario. Through these steps an application on physical hardware will likely be within reach.

\section*{Acknowledgments}
Marco Caccamo was supported by an Alexander von Humboldt Professorship endowed by the German Federal Ministry of Education and Research. Harald Bayerlein and David Gesbert are supported by the PERFUME project funded by the European Research Council (ERC) under the European Union's Horizon 2020 research and innovation program (grant agreement no. 670896).

\bibliography{biblio}
\bibliographystyle{ieeetr}

\end{document}